\documentclass[letterpaper, 10 pt, conference]{ieeeconf}  

\IEEEoverridecommandlockouts                              

\usepackage[centertags]{amsmath}
\usepackage{amssymb}

\usepackage{mathtools}
\newcommand{\defeq}{\vcentcolon=}

\DeclareMathOperator*{\argmax}{arg\,max}
\DeclareMathOperator*{\argmin}{arg\,min}

\usepackage{hyperref}

\usepackage{graphicx}
\usepackage[skip=8pt,font=footnotesize]{caption}
\usepackage{cite}

\usepackage{xcolor,colortbl}

\usepackage{color}
\definecolor{grey1}{RGB}{192,192,192}
\definecolor{grey2}{RGB}{178,178,178}
\definecolor{grey3}{RGB}{150,150,150}
\definecolor{grey4}{RGB}{119,119,119}
\definecolor{grey5}{RGB}{77,77,77}
\definecolor{green}{RGB}{112,173,71}
\definecolor{blue2}{RGB}{68,115,196}
\definecolor{red}{RGB}{192,0,0}
\definecolor{yellow}{RGB}{255,192,0}

\usepackage{balance}

\usepackage{algorithm}
\usepackage{algpseudocode}
\algnewcommand\algorithmicinput{\textbf{Input:}}
\algnewcommand\Input{\item[\algorithmicinput]}
\algnewcommand\algorithmicoutput{\textbf{Output:}}
\algnewcommand\Output{\item[\algorithmicoutput]}
\algnewcommand\algorithmicinit{\textbf{Initialize:}}
\algnewcommand\Init{\item[\algorithmicinit]}

\title{\LARGE \bf
Experience Selection Using Dynamics Similarity for Efficient Multi-Source Transfer Learning Between Robots
}

\author{Michael J. Sorocky\textsuperscript{*}, Siqi Zhou\textsuperscript{*}, and Angela P. Schoellig
\thanks{\textsuperscript{*}Equal contribution. The authors are with the Dynamic Systems Lab (\href{http://www.dynsyslab.org}{http://www.dynsyslab.org}), Institute for Aerospace Studies, University of Toronto, Canada. The authors are also affiliated with the Vector Institute for Artificial Intelligence, Toronto. Emails: msorocky@robotics.utias.utoronto.ca,  siqi.zhou@robotics.utias.utoronto.ca, schoellig@utias.utoronto.ca}%
}

\begin{document}

\maketitle
\thispagestyle{empty}
\pagestyle{empty}

\begin{abstract}
In the robotics literature, different knowledge transfer approaches have been proposed to leverage the experience from a source task or robot---real or virtual---to accelerate the learning process on a new task or robot. A commonly made but infrequently examined assumption is that incorporating experience from a source task or robot will be beneficial. In practice, inappropriate knowledge transfer can result in negative transfer or unsafe behaviour. In this work, inspired by a system gap metric from robust control theory, the $\nu$-gap, we present a data-efficient algorithm for estimating the similarity between pairs of robot systems. In a multi-source inter-robot transfer learning setup, we show that this similarity metric allows us to predict relative transfer performance and thus informatively select experiences from a source robot \textit{before} knowledge transfer. We demonstrate our approach with quadrotor experiments, where we transfer an inverse dynamics model from a real or virtual source quadrotor to enhance the tracking performance of a target quadrotor on arbitrary hand-drawn trajectories. We show that selecting experiences based on the proposed similarity metric effectively facilitates the learning of the target quadrotor, improving performance by 62\% compared to a poorly selected experience.

\end{abstract}


\section{Introduction}
\label{sec:introduction}
Knowledge transfer or transfer learning has been used in robotics to accelerate the learning process for acquiring new skills, or to further improve a robots' performance~(e.g.,~\cite{devin2017learning,stark2019experience}). At its core, transfer learning seeks to leverage existing experiences to initialize learning in a new domain that would otherwise be learned from scratch, thereby reducing the time and cost for training on physical or virtual robots. Examples of transfer learning in robotics include both
\textit{(i)} inter-task transfer learning, in which the experience learned for one task is used for learning new tasks on the same robot~\cite{fu2016one,stark2019experience,hamer2013knowledge}, and \textit{(ii)} inter-robot transfer learning, where the experience from one robot is used to accelerate or improve the learning on a second robot~\cite{bocsi2013alignment,ammar2015unsupervised,pereida2018data}.

While transfer learning has been demonstrated to improve robot performance, an implicit assumption that is less often examined is the similarity between the source and the target task or domain. As discussed in different transfer learning work~\cite{torrey2010transfer,taylor2009transfer,rosenstein2005transfer,wang2019characterizing}, transferring inappropriately selected experiences can deteriorate learning and lead to worse performance compared to not using any source data---this phenomenon is termed \textit{negative transfer}.  Approaches to avoid negative transfer include filtering out irrelevant source information~\cite{wang2019characterizing, rosenstein2005transfer}, or using data from multiple sources to increase the chance of positive transfer~\cite{yao2010boosting}. Although there exist approaches to encourage positive transfer and suppress negative transfer~\cite{torrey2010transfer}, negative transfer is usually detected only \textit{after} the transferred experience is applied to the target. This after-the-fact negative transfer detection can be inefficient and unsafe for robotic applications, particularly when physical robots are involved in the training process.

\begin{figure}
    \centering
    \includegraphics[trim = 0cm 0cm 0.6cm 11cm, clip, width=\columnwidth]{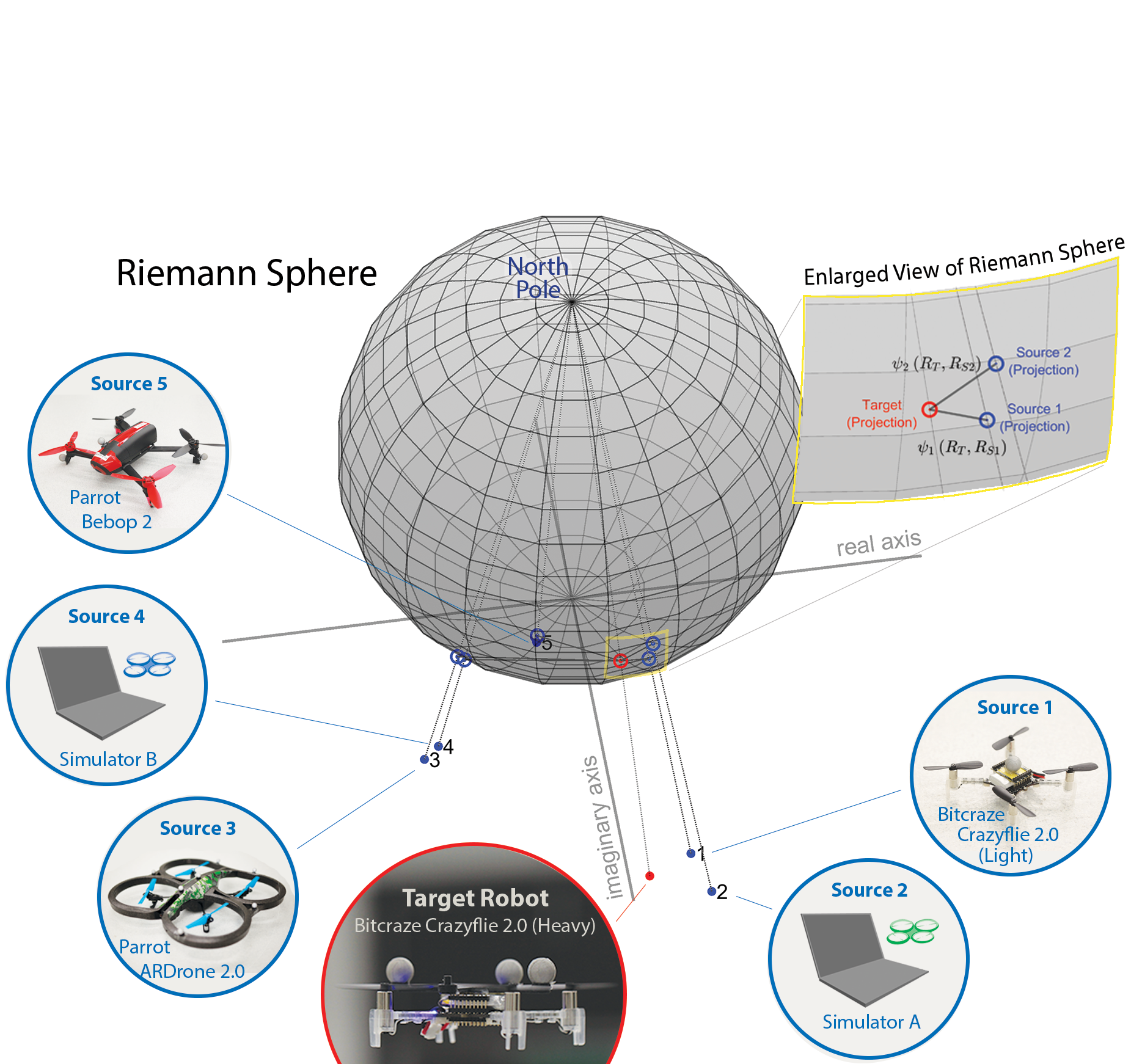}
    \caption{An illustration of the $\nu$-gap inspired dynamics similarity characterization between five target (red) and source (blue) robot pairs---points closer together on the sphere (open dots) are seen as more similar. In this work, we consider a multi-source inter-robot transfer learning setup, where we use the proposed similarity characterization to informatively select a source experience from a repository that best facilitates the learning process on the target robot. A video of this work can be found here: \url{http://tiny.cc/similarity}} 
    \label{fig:riemann_sphere}
    \vspace{-1.8em}
\end{figure}

As noted in~\cite{torrey2010transfer,rosenstein2005transfer}, the effectiveness of transfer learning typically depends on the similarity between the source and the target domain. Characterizing the similarity between the source and the target task or robot is thus crucial for understanding when transfer learning is safe and effective. In this paper, we consider a multi-source inter-robot transfer learning problem. Inspired by a notion of system gap from robust control theory, the $\nu$-gap, we propose a data-efficient algorithm for characterizing similarity between two robots with unknown dynamics (Fig.~\ref{fig:riemann_sphere}). We show that the proposed similarity characterization, computed with simple experiments on periodic trajectories, allows us to predict relative transfer performance and thus informatively select the source robot that best facilities learning on the target robot. 
Our contributions in this paper are as follows:
\begin{enumerate}
    \item Inspired by the $\nu$-gap metric from robust control theory~\cite{zhou1998essentials}, we propose a data-efficient approach for characterizing the similarity between two robot systems whose exact dynamics are not necessarily known.
    \item Based on this notion of similarity, we introduce an optimization framework that allows us to systematically evaluate experience from different source robots before transferring the experience to the target robot.
    \item We show in quadrotor experiments that an online learned inverse dynamics model with knowledge transferred from the most similar source quadrotor allows the target quadrotor to better track arbitrary hand-drawn trajectories in single attempts, as compared to transfer from less similar robots.
\end{enumerate}






\section{Related Work}
\label{sec:related_work}
    Transfer learning has been considered in a wide range of robot learning problems, which include, but are not limited to, robot model learning~\cite{bocsi2013alignment,helwa2017multi,makondo2015knowledge}, policy learning~\cite{stark2019experience,devin2017learning,gupta2017learning}, and objective function learning~\cite{pautrat2018bayesian,marco2017virtual}. Whether it is inter-task transfer~\cite{pautrat2018bayesian,stark2019experience}, inter-robot transfer~\cite{bocsi2013alignment,helwa2017multi,makondo2015knowledge,gupta2017learning}, or transferring experience from simulation to reality~\cite{cutler2015efficient,marco2017virtual}, the common goal in transfer learning is to leverage existing experiences to accelerate or improve the learning process of the target task or robot. 
    
    While in an inter-task transfer framework there exist discussions on characterizing task similarity for multi-source experience selection via a maximum likelihood model selection approach~\cite{pautrat2018bayesian}, or based on the difference of task descriptor representations~\cite{stark2019experience}, discussions on a similarity characterization of robot dynamics for inter-robot transfer problems are rare. As similarity and transfer performance are positively correlated concepts~\cite{torrey2010transfer,rosenstein2005transfer}, in inter-robot transfer problems, especially those involving robot model learning, characterizing the \textit{dynamics similarity} of the robots is essential for predicting the effectiveness of transferring an experience. In this work, we consider a multi-source transfer learning setup similar to~\cite{pautrat2018bayesian,stark2019experience}, but we focus on an inter-robot model transfer problem, where our goal is to select the best source robot to facilitate the learning of a target robot.
    
    A concept that can be used for quantifying the distance between two dynamical systems is the notion of system gap from robust control theory~\cite{zhou1998essentials}. In robust control, the incentive of characterizing system similarity is to provide theoretical guarantees in the presence of model uncertainties. For instance, for a classical feedback control architecture, it has been shown that an uncertain system can be stabilized by a controller designed based on a model that is sufficiently close in the sense of system gap~\cite{zhou1998essentials,zames1981uncertainty}. In addition to providing stability guarantees, gap metrics have also been applied to multi-linear model control, where a set of linear system models are used to approximate the dynamics of a nonlinear system. The control input to the nonlinear system is comprised of adding control inputs computed based on each linear system, each weighted with the gap metric between the linear system and a linearization of the nonlinear system at the current operating point~\cite{arslan2004multi,du2014gap}. 
    
    Although gap metrics have been shown as an appropriate measure of dynamics similarity between systems, the computation of the gap metrics requires either detailed system models~\cite{zhou1998essentials,georgiou1988computation} or sufficiently large amounts of data~\cite{koenings2017data}, which can be challenging or non-economical to obtain for practical robot systems. In this work, inspired by the notion of system gap~\cite{zhou1998essentials}, we propose a data-efficient optimization algorithm for quantifying the similarity between two robots whose dynamics are unknown. We further leverage this similarity characterization in an inter-robot transfer learning problem, and show that the proposed similarity characterization allows us to effectively select the best source robot for improving the performance of the target robot.
\section{Problem Formulation}
\label{sec:problem_formulation}
We consider a multi-source inter-robot transfer learning problem, where a target robot improves its performance through learning and has access to experience from a set of $N$ source robots, which can be leveraged to facilitate its own learning. Our goal is to develop an algorithm that allows us to quantify the similarity between each source and the target robot such that we can efficiently select experiences from a source robot that best facilitates the learning on the target robot. We aim to show that selecting an experience from a source robot similar to the target robot in the sense of the proposed metric will result in better transfer performance.

\begin{figure}[t]
	\vspace{1.2ex}
	\centering
	\includegraphics[trim={0cm 0.5cm 0cm 0cm},width=\columnwidth]{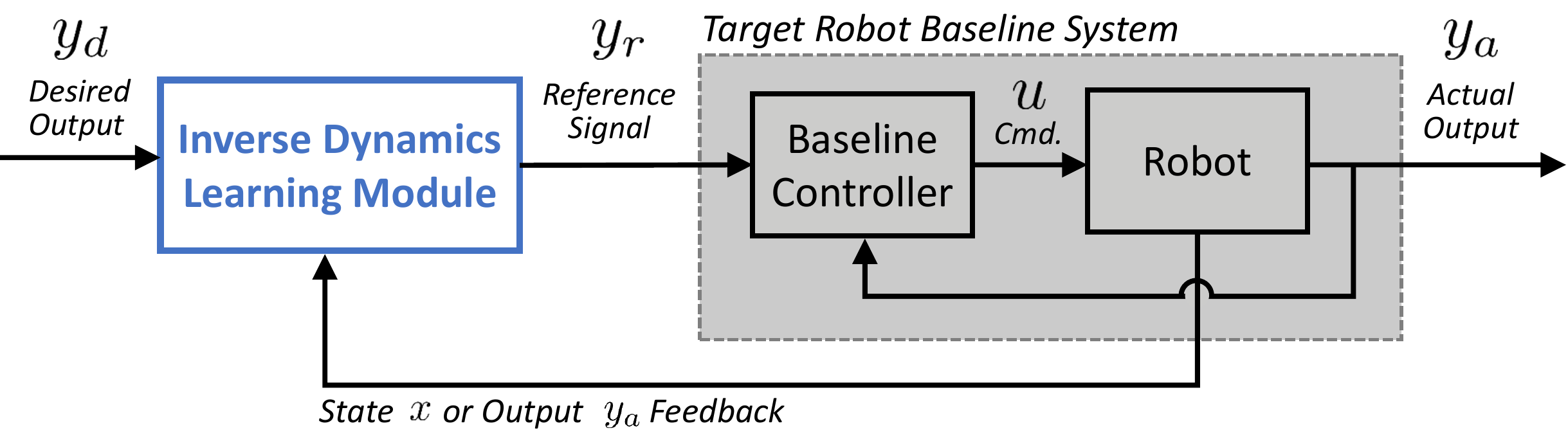}
	\vspace{-1em}
	\caption{Block diagram of an inverse learning approach for enhancing the tracking performance of black-box systems on arbitrary trajectories~\cite{li-icra17,zhou-cdc17}. The inverse dynamics module (blue box) represents the inverse of the target robot baseline system (grey box); it is pre-cascaded to the target robot baseline system to adjusts the reference signal $y_r$ to reduce the trajectory tracking error of the baseline system (i.e., $e = y_d - y_a$, where $y_d$ denotes the desired output and $y_a$ denotes the actual output of the target robot system). The inverse dynamics module is implemented as a GP and the inverse dynamics model of a source system is used as the prior mean function to facilitate online learning of the target inverse dynamics (Eqn.~\eqref{eqn:gp_inverse_model}).}
	\label{fig:block-diagram}
    \vspace{-1.5em}
\end{figure}

To analyze the proposed similarity characterization, we consider an inverse dynamics learning framework shown in Fig.~\ref{fig:block-diagram}. Through leveraging the inverse dynamics from a source robot, we aim to enhance the impromptu tracking performance of the target robot on arbitrary hand-drawn trajectories using \textit{minimal online data} from the target robot. 

We denote the input and output of the inverse dynamics module by $\xi$ and $\gamma$, and assume the inverse dynamics of the target robot are represented by a Gaussian Process (GP),
\begin{equation}
    F_T(\xi) \sim\mathcal{GP}\left(\mu(\xi), \kappa(\xi,\xi') \right),
\end{equation}
where $\mu(\cdot)$ and $\kappa(\cdot,\cdot)$ are the prior mean and covariance function of the GP, respectively. 
To facilitate the training of the GP inverse dynamics model of the target robot, we leverage the inverse dynamics model of a previously trained source robot as the prior mean function. The posterior mean $f_T(\xi)$ and variance $\sigma_T^2(\xi)$ of the GP at a new input $\xi_*$ are
\begin{equation}
    \begin{aligned}
        f_T(\xi_*) &= f_{S_n}(\xi_*) + \mathbf{k}^T(\xi_*)\left(\mathbf{K}+\sigma_n^2I \right)^{-1}(\boldsymbol{\gamma} - \mathbf{f}_{S_n})\\
        \sigma_T^2(\xi_*) &= \kappa(\xi_*,\xi_*) - \mathbf{k}^T(\xi_*)\left(\mathbf{K}+\sigma_n^2I \right)^{-1}\mathbf{k}(\xi_*),
    \end{aligned}
    \label{eqn:gp_inverse_model}
\end{equation}
where $f_{S_n}(\cdot)$ is a deterministic inverse dynamics model of source robot~$n$, $\boldsymbol{\gamma} = [\gamma_1,\dots,\gamma_D]^T$ and $\mathbf{f}_{S_n} = [f_{S_n}(\xi_1),\dots,f_{S_n}(\xi_D)]^T$ with $\{\xi_d, \gamma_d\}_{d=1}^{d=D}$ being the input-output data collected online from the target robot, $\mathbf{k}(\xi_*)\defeq [\kappa(\xi_*,\xi_1),\dots,\kappa(\xi_*,\xi_D)]^T$, $\mathbf{K}\in\mathbb{R}^{D \times D}$ is the covariance matrix with entries $[\mathbf{K}]_{ij}=\kappa(\xi_i,\xi_j)$, $\sigma_n^2$ is the noise variance, and $I\in\mathbb{R}^{D \times D}$ is the identity matrix.

Following the discussion in~\cite{zhou-cdc17}, for the inverse dynamics learning approach to be safely applied, we assume that the target and the source robot systems are minimum-phase (i.e., are stable and have stable inverse dynamics). This assumption is not restrictive and is satisfied by practical robot systems such as quadrotors~\cite{zhou-cdc17} and robot manipulators~\cite{spong2006modeling}.

\section{Experience Selection Through a Characterization of Robot Dynamics Similarity}
\label{sec:similarity-characterization}
In this section, we outline an algorithm for prior mean selection based on a characterization of similarity between robot dynamics. To facilitate our discussion, in Sec.~\ref{sub:background_gap}, we first present a brief background on the $\nu$-gap metric from robust control theory. Inspired by the $\nu$-gap metric, we formulate an optimization problem based on a characterization of system similarity for experience selection in Sec.~\ref{subsec:opt-problem}. In Sec.~\ref{subsec:proposed-solution}, we propose a Bayesian Optimization (BO)-based algorithm with a novel acquisition function to solve this optimization problem for robots with unknown dynamics.

\subsection{Background on the $\nu$-Gap Metric}
\label{sub:background_gap}

One notion of system similarity is the $\nu$-gap metric, which provides a measure of distance between two linear dynamical systems \cite{vinnicombe1993frequency}. In this subsection, we present the essential concepts to facilitate our discussion. An in-depth analysis of the metric can be found in \cite{vinnicombe1993frequency, zhou1998essentials,vinnicombe2001uncertainty}.

For linear dynamical systems, the response to a periodic input $u$ is a periodic output $y$ with the same frequency but with different magnitude and phase. Given a periodic input with frequency $\omega$ to a dynamical system, we define a complex number $z(\omega)= P(j\omega) = A(\omega)\mathrm{e}^{j\theta(\omega)}$ to characterize the input-output response at the input frequency~$\omega$, where $P(s)$ is the transfer function of the system, $A(\omega)$ and $\theta(\omega)$ are the amplitude gain and phase shift in the input-output response, and $j$ is the imaginary unit.


In \cite{zhou1998essentials}, it is shown that for two minimum phase systems $P_1,\ P_2$, under the condition that $||P_1(-j\omega)P_2(j\omega)||_\infty<1$, the $\nu$-gap metric can be computed by
\begin{equation}
    \delta_\nu(z_1, z_2) \defeq \sup_{\omega\in\mathbb{R}} \psi(z_1(\omega), z_2(\omega)),
    \label{eq:vgap}
\end{equation}
where $z_1,\ z_2$ characterize the input-output response of $P_1,\ P_2$ for an input frequency $\omega$,  and
\begin{equation}
    \psi(z_1(\omega), z_2(\omega)) = \frac{|z_1(\omega)-z_2(\omega)|}{\sqrt{1+ |z_1(\omega)|^2}\sqrt{1+ |z_2(\omega)|^2}}.
	\label{eq:psi}
\end{equation} 

The function $\psi$ in Eqn. \ref{eq:psi} can be given a geometric interpretation as the Euclidean distance between the projection of two points $z_1(\omega),\ z_2(\omega)\in\mathbb{C}$ onto the Riemann sphere, a sphere of radius $0.5$ tangent to the origin of the complex plane. The solid red and blue dots in Fig.~\ref{fig:riemann_sphere} are projected onto the sphere by connecting a line between each point and the point $(0,0,1)$ (the \textit{North Pole}) on the sphere --- the intersection of these lines and the sphere (the open red/blue dots in Fig. \ref{fig:riemann_sphere}) are their respective projections. The function $\psi$ gives the Euclidean distance between the projected points (open red and blue dots in Fig.~\ref{fig:riemann_sphere}) --- this distance is termed the \textit{chordal distance}. The $\nu$-gap metric represents the largest chordal distance over all frequencies $\omega$. It can be shown that $\delta_\nu$ satisfies $0\leq \delta_\nu(z_1,z_2)\leq1$~\cite{zhou1998essentials}. We use the $\nu$-gap metric as a measure of similarity, interpreting values of $\delta_\nu$ closer to~$0$ meaning the two systems are more similar. 

\subsection{Experience Selection as an Optimization Problem}
\label{subsec:opt-problem}
Given a set of $N$ source robots $\mathcal{S}$, we aim to select an experience from the source robot with the most similar dynamics to the target to improve the performance of the target robot. We denote the $n$th source robot by $R_{S_n}$, and the target robot by $R_T$. The gain $A_R$ and phase shift $\theta_R$ of a robot's response to an input with frequency $\omega$ are characterized by
\begin{equation}
\label{eqn:z_r}
    z_R(\omega)=A_R(\omega)\mathrm{e}^{j\theta_R(\omega)},
\end{equation}
where $R$ denotes the target robot $R_T$ or a source robot $R_{S_n}$.

Inspired by the $\nu$-gap metric, we propose the following characterization of dynamics similarity between a source and target robot over a set of operating frequencies of interest $\mathcal{W} \defeq [\omega_{\text{min}}, \omega_{\text{max}}]$:
\begin{equation}
    \psi_n^* \defeq \max_{\omega \in \mathcal{W}} \psi_n(\omega),
    \label{eq:similarity-characterization}
\end{equation}
where using Eqn.~\eqref{eq:psi} we define
\begin{equation}
    \psi_n(\omega) \defeq \psi(z_T(\omega), z_{S_n}(\omega)).
    \label{eq:psi-n}
\end{equation}
Analogous to the $\nu$-gap metric, it holds that $0\leq \psi_n^* \leq 1$, and we interpret values of $\psi_n^*$ close to 0 indicating the two systems are similar. In contrast to $\delta_\nu(\cdot,\cdot)$ in Eqn.~\eqref{eq:vgap}, we consider a finite frequency range of interest $\mathcal{W}$. 

With this notion of dynamics similarity, we formulate experience selection as the following optimization problem:
\begin{equation}
    R_S^* \defeq \argmin_{n}\ \psi_n^*,
    \label{eq:min-max-psi}
\end{equation}
where $\psi_n^*$ is the characterization of dynamics similarity in Eqns.~\eqref{eq:similarity-characterization} and \eqref{eq:psi-n}. Intuitively, the optimization problem in Eqn.~\eqref{eq:min-max-psi} selects the source system $R_S^*$ whose maximum chordal distance (i.e., the worst-case distance) from the target system over the range $\mathcal{W}$ is the minimal. Given the inverse dynamics learning framework in Fig.~\ref{fig:block-diagram}, by considering the worst-case chordal distance between the target and the source robot systems, we select a source inverse dynamics model that facilitates high-performance tracking for arbitrary trajectories over the range $\mathcal{W}$.

Note that in robust control, system similarity metrics are often computed based on detailed models of the systems. In the next subsection, we outline a data-efficient algorithm to estimate dynamics similarity and select an experience based on Eqn.~\eqref{eq:min-max-psi} for robots with possibly unknown dynamics.

\subsection{Experience Selection for Source and Target Robot Systems with Unknown Dynamics}
\label{subsec:proposed-solution}

We first outline an approach to estimate the similarity metric in Eqn.~\eqref{eq:similarity-characterization} using data from a source robot and a target robot, then extend the solution to the multi-source experience selection problem in Eqn.~\eqref{eq:min-max-psi}. To solve the experience selection problem, we first note that while the objective function $\psi_n(\omega)$ is unknown, we can sample from it at selected frequencies $\omega_{\text{sample}}$. In particular, given a frequency $\omega_{\text{sample}}$, we can send a periodic input to the source and the target robots to estimate the amplitude gain $A(\omega_{\text{sample}})$ and phase shift $\theta(\omega_{\text{sample}})$ from input-output data. This information allows us to calculate $z_T(\omega_{\text{sample}})$ and $z_{S_n}(\omega_{\text{sample}})$ defined in Eqn.~\eqref{eqn:z_r}, and $\psi_n(\omega_{\text{sample}})$ defined in Eqn.~\eqref{eq:psi-n}.


For solving the problem in Eqn.~\eqref{eq:similarity-characterization}, we propose a Bayesian Optimization (BO)-based algorithm to find the maximum of $\psi_n$. 
In particular, we model $\psi_n$ by a GP and find its maximum by iteratively sampling the objective function based on an acquisition function. We denote the GP posterior mean and variance by $\hat{\psi}_n(\omega)$ and $\hat{\sigma}_n^2(\omega)$. Given $M$ previous samples $\{\psi_n(\omega_m)\}_{m=1}^M$, we define $\psi_{n,\text{max}} \defeq \max_{k} \psi_n(\omega_k)$. A common BO acquisition function is the expected improvement (EI), which is given for $\omega\in\mathcal{W}$ by
\begin{equation}
     \text{EI}_n(\omega) = 
        (\hat{\psi}_n(\omega) - \psi_{n,\text{max}}-\xi)\Phi(Z) + \hat{\sigma}_n(\omega)\phi(Z)
    \label{eq:ei}
\end{equation}
if $\hat{\sigma}_n(\omega)\neq0$, and 0 otherwise, where $Z \defeq \frac{\hat{\psi}_n(\omega) - \psi_{n,\text{max}}-\xi}{\hat{\sigma}_n(\omega)}$, $\Phi$ and $\phi$ are the cumulative and probability density functions of the standard normal distribution, and $\xi$ is an exploration parameter~\cite{shahriari2015taking}.

The maximum of $\psi_n$ is found through an iterative process: we sample $\psi_n$ at $\omega_{\text{sample}}\defeq \argmax_{\omega \in \mathcal{W}} EI(\omega)$, we refit the GP with the new sample, and the process is repeated. Upon convergence, the estimate of similarity $\hat{\psi}_n^*$ between the target robot and the $n$th source robot is given by the maximum of the posterior mean of its GP approximation:
\begin{equation}
    \hat{\psi}_n^* \defeq \max_{\omega \in \mathcal{W}}\hat{\psi}_n(\omega),
    \label{eq:quantify-similarity}
\end{equation}
with a smaller $\hat{\psi}_n^*$ representing a more similar source robot.

For solving the experience selection problem in Eqn.~\eqref{eq:min-max-psi}, we use a GP to model the unknown chordal distance function $\psi_n$ for each source and target robot pair. To account for the outer-loop minimization problem in Eqn. \eqref{eq:min-max-psi}, we modify the acquisition function as follows:
\begin{equation}
    \alpha(\omega) \defeq \max_{R_{S_n} \in \mathcal{S}} \text{EI}_n(\omega)/\hat{\psi}_n^*,
    \label{eq:acquisition}
\end{equation}
where $\text{EI}_n(\omega)$ is the expected improvement of the point $\omega$ for the $n$-th GP, and $\hat{\psi}_n^*$ is the similarity estimate for $R_{S_n}$. 
Intuitively, the acquisition function in Eqn. \eqref{eq:acquisition} encourages the selection of a sample $\omega_{\text{sample}}$ at the worst-case chordal distance for more similar robots.

Based on this BO framework, we outline our proposed algorithm to determine system similarity for experience selection in Alg. \ref{alg:gap_estimation}. The BO algorithm is initialized with a uniformly sampled $\omega_\text{sample}$. We sample from each GP by running periodic trajectories on each source and target system until convergence. Given the estimates of similarity between each source robot and the target robot, the most similar robot $R^*_S$ has minimum $\hat{\psi}_n^*$: $R^*_S = \argmin_n \hat{\psi}_n^*$, solving the experience selection problem posed in Eqn.~\eqref{eq:min-max-psi}.

\setlength{\textfloatsep}{2pt}

\begin{algorithm}[t]
\vspace{0.5em}
\centering
\begin{algorithmic}[1]
\caption{Source/Target Robot Similarity Estimation}
\label{alg:gap_estimation}
\Input One target robot, $N$ source robots and a frequency operating range $\mathcal{W}$
\Output A similarity estimate of each source robot to the target, $\hat{\psi}_n^*$, and the source robot $R_S^*$ most similar to the target over $\mathcal{W}$
\Init Empty datasets: $\mathcal{D}_{S_n} \leftarrow \emptyset\ (n=1,\ \dots,\ N)$
\item Compute initial sample location: $\omega_{\text{sample}} \sim \mathcal{U}(\omega_{\text{min}}, \omega_{\text{max}})$
\While {not converged} 
    \State Estimate $z_T(\omega_{\text{sample}})$ \label{algline:estimate-zt}
    \For {$n=1,\ \dots,\ N$}
        \State Estimate $z_{S_n}(\omega_{\text{sample}})$ \label{algline:estimate-zs}
        \State Compute $\psi_n(\omega_{\text{sample}})$ (Eqn. \eqref{eq:psi-n}) \label{algline:compute-psi}
        \State $\mathcal{D}_{S_n} \leftarrow \mathcal{D}_{S_n} \cup \{(\omega_{\text{sample}}, \psi_n(\omega_{\text{sample}})\}$
        \State Fit $n$th GP with new data $\mathcal{D}_{S_n}$ \label{algline:fit-gp}
    \EndFor
    \State Compute $\omega_{\text{sample}} = \argmax_{\omega \in \mathcal{W}} \alpha(\omega)$
\EndWhile
\State $\hat{\psi}_n^* \defeq \max_\omega \hat{\psi}_n(\omega)$ (Eqn. \eqref{eq:quantify-similarity}), and $R^*_S = \argmin_n \hat{\psi}_n^*$
\end{algorithmic}
\end{algorithm}

\section{Simulation}
\label{sec:simulation_results}
We first illustrate the proposed algorithm~(Alg.~\ref{alg:gap_estimation}) for characterizing dynamics similarity between robots in simulation. 

\subsection{Simulation Setup}
We consider a target robot and a set of three source robots whose dynamics are represented by 
\begin{equation}
    \begin{aligned}
        \mathbf{x}(k+1) &= \begin{bmatrix}
        1&T_s-\beta\\0&1-\beta
        \end{bmatrix}\mathbf{x}(k) + \alpha \begin{bmatrix}\frac{1}{2}T_s^2\\T_s
        \end{bmatrix}u(k)\\
        y_a(k) &= \begin{bmatrix}1 & 0\end{bmatrix}\mathbf{x}(k),
    \end{aligned}
\end{equation}
where $k\in\mathbb{Z}_{\geq0}$ is the discrete-time index, $\mathbf{x}\in\mathbb{R}^2$ is the state, $y_a\in\mathbb{R}$ is the output, $u\in\mathbb{R}$ is the command sent to the robot, $T_s=0.015$ sec is the sampling time, and $(\alpha, \beta)$ are two system parameters that differentiate the source and the target robot dynamics. In this simulation, the parameters  for the target robot are $(0.85, 0.003)$; for the three source robots, the parameters are $(1.0, 0.003)$, $(0.97, 0.004)$, and $(0.9, 0.001)$, respectively. As shown in Fig.~\ref{fig:block-diagram}, we aim to improve a baseline tracking controller using online learning. For illustration purposes, we use a simple PD baseline tracking controller defined as follows: $u(k) = k_pe(k) + k_d \dot{e}(k)$, where $k_p = 5$ and $k_d = 4.5$ are the baseline controller parameters and $e(k) = y_r(k) - y_a(k)$. Note that, when applying the proposed algorithm, we assume that the dynamics of the source and the target robot baseline systems are unknown: we only use the input and output data of the source and the target robots from frequency tests.

Motivated by the inverse dynamics learning application (Fig.~\ref{fig:block-diagram}), we are interested in characterizing the similarity between the source and the target robot baseline systems (i.e., the dynamics from $y_r$ to $y_a$). In this study, we apply Alg.~\ref{alg:gap_estimation} to estimate the similarity of the three target/source robot pairs. For the simulation, we set $\mathcal{W}=[0,10]$~rad/sec, and the algorithm is initialized with a randomly generated input frequency $\omega_\text{sample}$. In each iteration, we send a periodic reference trajectory $y_r$ with frequency $\omega_{\text{sample}}$ to the source and target robots. Based on the input-output response, we estimate $z_T$ and $z_{S_n}$ to calculate $\psi_n(\omega_{\text{sample}})$ in Eqn.~\eqref{eq:psi-n}. For each source robot, iteratively augmented data $\mathcal{D}_{S_n}$ is used to fit a GP model and estimate the similarity $\hat{\psi}_n^*$ of the source and target robot pair. We use GPs with a zero prior mean function and the Mat\'ern 3/2 kernel; the kernel hyperparameters are optimized in each iteration to maximize the marginal likelihood of the GP model. As formulated in Eqn.~\eqref{eq:min-max-psi}, the `best' source robot is the one that minimizes the worst-case chordal distance, $\max_{\omega\in \mathcal{W}}\psi_n(\omega)$.
\begin{figure}
    \centering
    \includegraphics[trim={0.6cm 1.45cm 1cm 0cm}, width=\columnwidth]{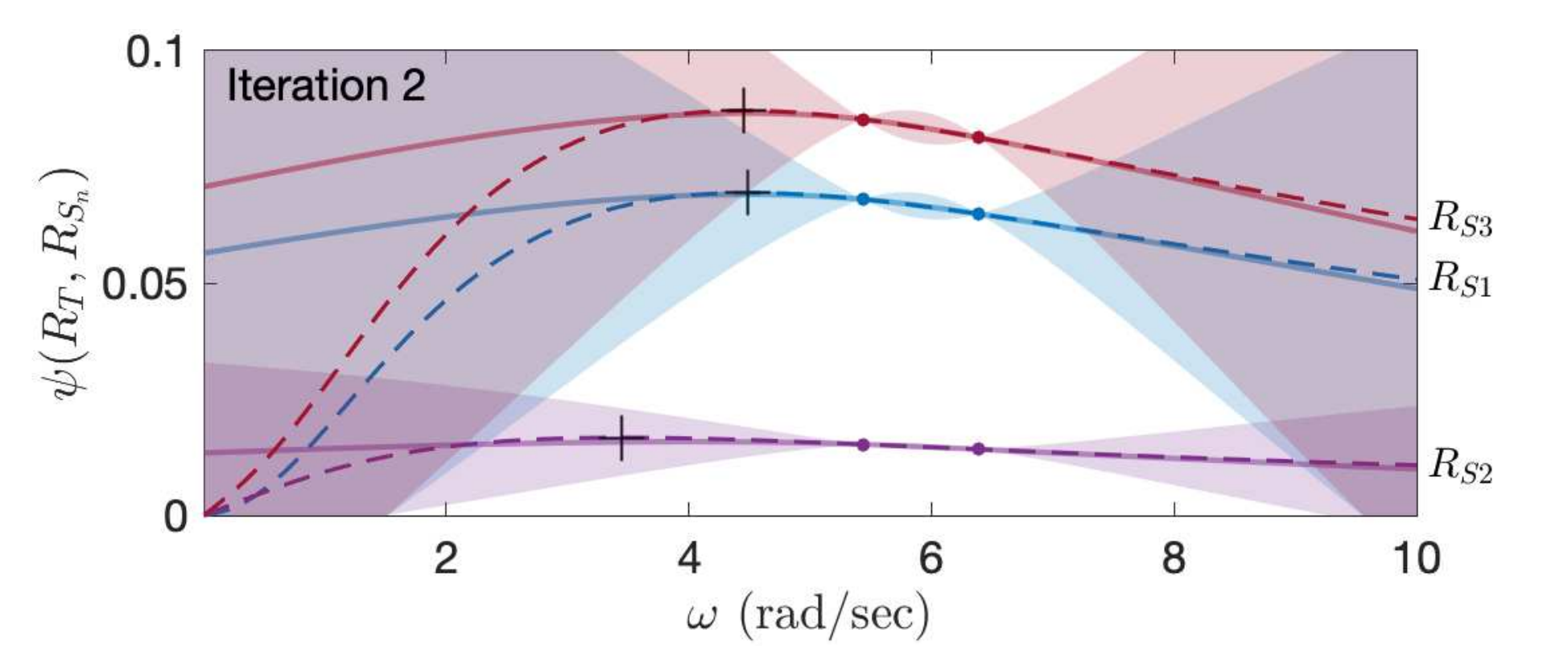}
    \includegraphics[trim={0.6cm 0.3cm 1cm 0cm}, width=\columnwidth]{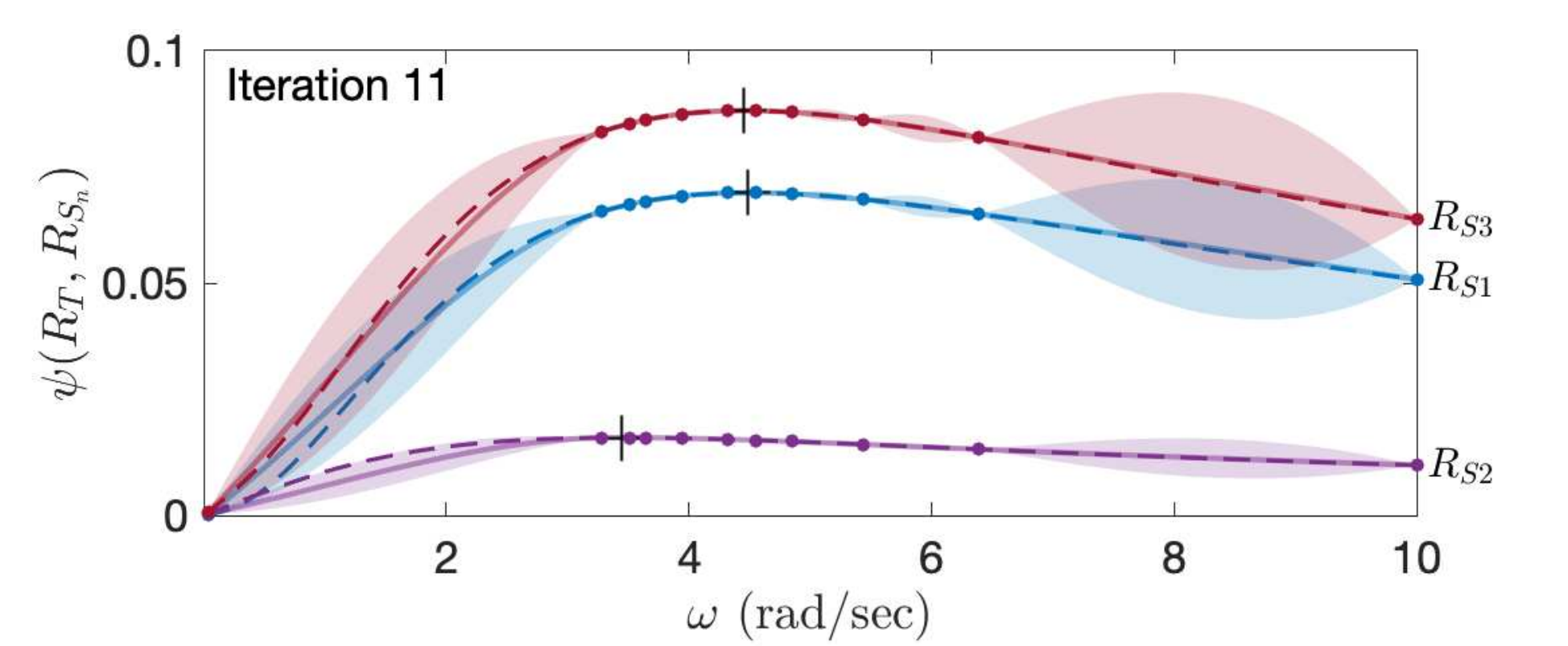}
    \caption{Illustration of the proposed dynamics similarity characterization outlined in Alg.~\ref{alg:gap_estimation}. The top and bottom panels respectively show the similarity estimation in an earlier and a later iteration of the algorithm. The true chordal distance $\psi_n(\omega)$ between a source and a target pair and its GP approximation $\hat{\psi}_n(\omega)$ are represented by the dashed line and solid line, respectively. The maximum of $\psi_n(\omega)$ for each source and target robot pair is indicated by `$+$', and the sampled training points are indicated by `$\bullet$'.}
    \label{fig:bo_sim}
\end{figure}
\subsection{Simulation Results} Figure~\ref{fig:bo_sim} shows the true chordal distance $\psi_n$ (dashed line), and its GP approximation $\hat{\psi}_n$ (solid line) for each source/target pair in an earlier and a later BO iteration. 
As can be seen from Fig.~\ref{fig:bo_sim}, the proposed algorithm encourages sampling around the maxima of the chordal distance curves for each source/target pair. This allows us to effectively estimate the worst-case distance between the source and the target robot and quantify their similarity. The similarity estimation converges to the maxima of the true function $\psi_n$ (indicated by `$+$') in seven iterations. In contrast to traditional approaches that require either an fully identified model or sufficient excitation of the robots, we see that the proposed algorithm requires minimal data from a few simple frequency tests. The converged similarity estimates $\hat{\psi}_n^*$ are 0.069, 0.017, and 0.087, respectively, for the three source/target robot pairs, respectively, with $R_S^* = R_{S_2}$. In the next section, we demonstrate the similarity characterization for experience selection in quadrotor tracking experiments.
\section{Quadrotor Experiments}
\label{sec:quadrotor_experiments}
In this section, we demonstrate the experience selection algorithm from Sec. \ref{subsec:proposed-solution} with quadrotor experiments. 

\subsection{Experiment Setup}
\label{subsec:quad-experiment-setup}
The goal of the experiments is to improve the tracking performance for a target quadrotor $R_{T}$ Bitcraze Crazyflie 2.0 (mass 36g) over its feasible operating frequencies $\mathcal{W} = [0.1,\ 5.4]$~rad/sec. We consider five source quadrotors shown in Fig.~\ref{fig:riemann_sphere}: ($R_{S_1}$) Bitcraze Crazyflie 2.0 (mass 31g), ($R_{S_2}$) Crazyflie Simulator, ($R_{S_3}$) Parrot ARDrone 2.0, ($R_{S_4}$) ARDrone Simulator, and ($R_{S_5}$) Parrot Bebop 2. They have different baseline tracking controller designs and physical properties. The mass of source quadrotors ranges between 0.86--14.5x that of the target, and the rotor-to-rotor distance of the source quadrotors ranges between 1--3.9x that of the target.
As in our earlier work~\cite{zhou-cdc17,li-icra17}, each source quadrotor has a DNN inverse dynamics model previously trained on a rich dataset consisting of 2100 data points. The DNNs of the source quadrotors are fully-connected networks with four hidden layers of 128 ReLU neurons. To facilitate online learning of the target inverse dynamics model (Fig.~\ref{fig:block-diagram}), we are interested in characterizing the similarity between the source and the target robot baseline systems to select the source quadrotor with the most similar closed-loop dynamics. 

\begin{figure}
	\centering
	\includegraphics[trim={0cm 0.3cm 0cm 0cm},width=\columnwidth]{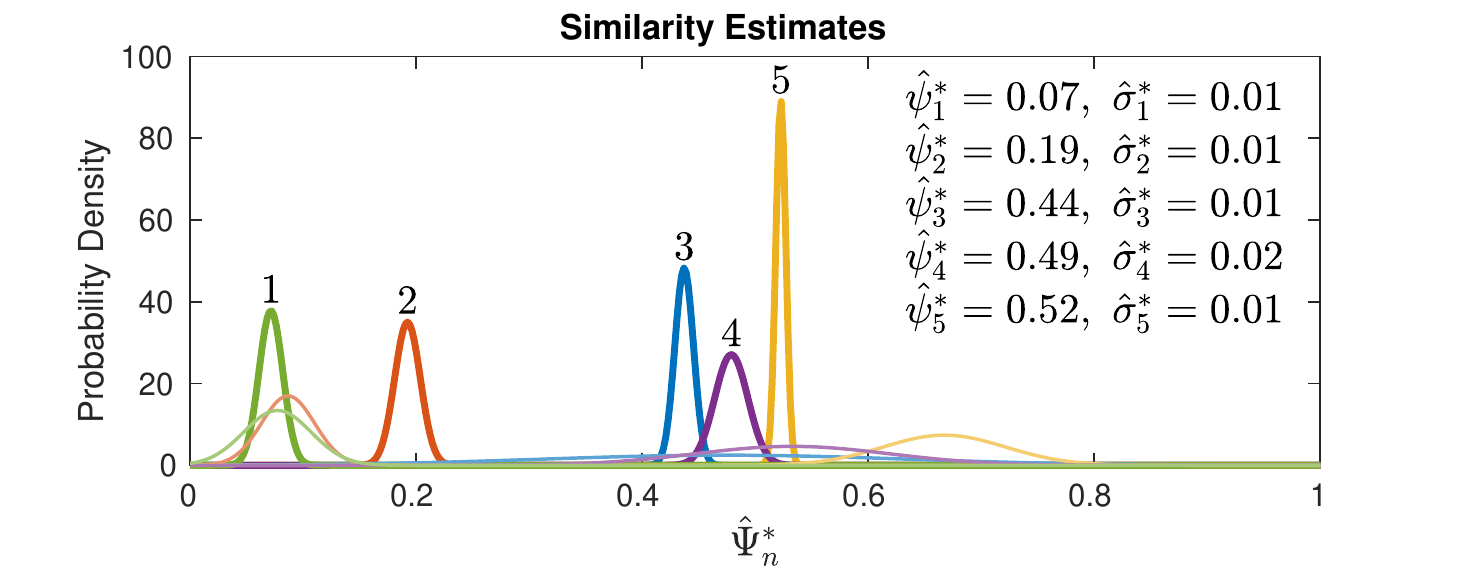}
	\caption{Similarity estimates between the target quadrotor and the five real/virtual source quadrotors: $\hat{\Psi}^*_n\sim\mathcal{N}(\hat{\psi}_n^*, \hat{\sigma}^{*^2}_n)$, where $\hat{\psi}_n^*$ is the similarity estimate defined in Eqn.~\eqref{eq:quantify-similarity} and $\hat{\sigma}^{*^2}_n$ is the corresponding variance. The thin lines correspond to estimates at the second iteration of Alg.~\ref{alg:gap_estimation}, while the thick lines correspond to their converged distributions in the ninth (final) iteration. Initially, sources 1 and 2 are equally similar to the target, while estimates for sources 3--5 are very uncertain. At the final iteration, the estimates are more certain that sources 1 and 2 are more similar to the target robot, in contrast to sources 3--5 which are less similar. The first source robot is most similar to the target (i.e., $R_S^* = R_{S_1}$).}
	\label{fig:gap-plot-experiments}
\end{figure}

\subsection{Quadrotor Experience Selection}
\label{subsec:quad-experience-selection}
In this subsection, we present the results of applying Alg.~\ref{alg:gap_estimation} to the inter-quadrotor transfer problem. To estimate $z_T(\omega_{\text{sample}})$ and $z_{S_n}(\omega_{\text{sample}})$, we run a sinusoidal trajectory of frequency $\omega_{\text{sample}}$ along the $x$-axis on each target and source system, using their respective \textit{baseline} tracking controllers (i.e., we do \textit{not} use any inverse modules in this step). We assume that the response along the $y$-axis of each quadrotor is similar to their response along the $x$-axis. We estimate the magnitude $A(\omega_{\text{sample}})$ and phase $\theta(\omega_{\text{sample}})$ for each quadrotor from its input-output response data. 

We ran Alg.~\ref{alg:gap_estimation} for nine iterations and observed that the estimated similarity $\hat{\psi}_n^*$ between the source and the target robot systems converged to approximately constant values. Fig.~\ref{fig:gap-plot-experiments} shows the posterior distributions $\hat{\Psi}^*_n\sim\mathcal{N}(\hat{\psi}_n^*, \hat{\sigma}^{*^2}_n)$ of the estimated similarity in the second and the final iteration of the algorithm, where $\hat{\psi}_n^*$ is the similarity estimate defined in Eqn.~\eqref{eq:quantify-similarity}, and $\hat{\sigma}^{*^2}_n$ is the corresponding variance. We observe the progression of the similarity estimates from the second to the final iteration of Alg.~\ref{alg:gap_estimation}. 
As seen in Fig.~\ref{fig:gap-plot-experiments}, sources 1 and 2 are more similar to the target robot and sources 3--5 are less similar. The most similar robot is $R_{S_1}$, and is expected to best facilitate the target quadrotor inverse learning. As we verify in the next set of experiments, the proposed experience selection algorithm allows us to infer the \textit{relative} performance of transferring experiences from different source robots \textit{before} testing on the target robot.
\subsection{Inverse Dynamics Learning and Tracking Experiments}
\label{subsec:experimental-results}
In this subsection, we verify the efficacy of the experience selection algorithm in the framework of Fig.~\ref{fig:block-diagram}. Specifically, we validate that using an inverse dynamics module from a more similar robot yields improved performance compared to using a module from a less similar robot. We modify the reference $y_r$ to the target baseline system using a GP inverse dynamics model trained on small amounts of online data from the target robot. The GP has a prior mean function given by a previously trained source DNN inverse dynamics model, and a Mat\'ern 3/2 kernel function. The kernel hyperparameters are initialized for all source robots as $\sigma_n^2 = 0.002$ (measurement noise), $\sigma_f^2 = 0.1$ (prior variance), and $\ell = 0.5$ (lengthscale)~\cite{rasmussen2006gaussian}. We implement the online module using the Python GPy library, and optimize hyperparameters online by maximizing the marginal likelihood of the GP. 

\begin{figure}
    \centering
    \includegraphics[width=\columnwidth]{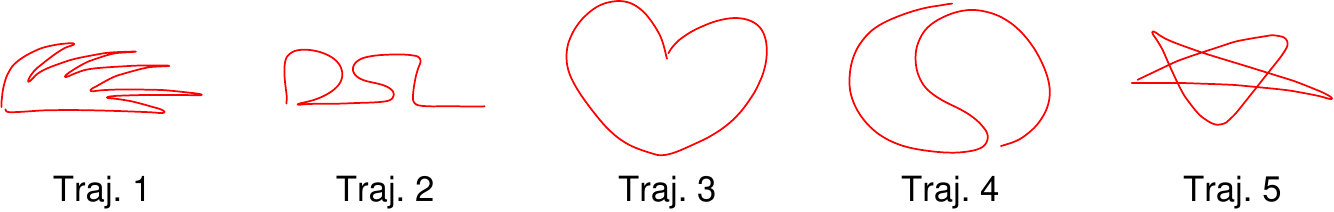}
    \caption{Five hand drawings for generating desired trajectories ($y_d$) in the quadrotor experiments. Trajectories are scaled based on three maximum speeds to test the performance for different aggressiveness of maneuvers.}
    \label{fig:desired_trajectory}
    \vspace{-0.5em}
\end{figure}

\begin{figure}[t]
	\centering
	\includegraphics[trim={2.5cm 2.5cm 7.0cm 1cm}, clip, width=\columnwidth]{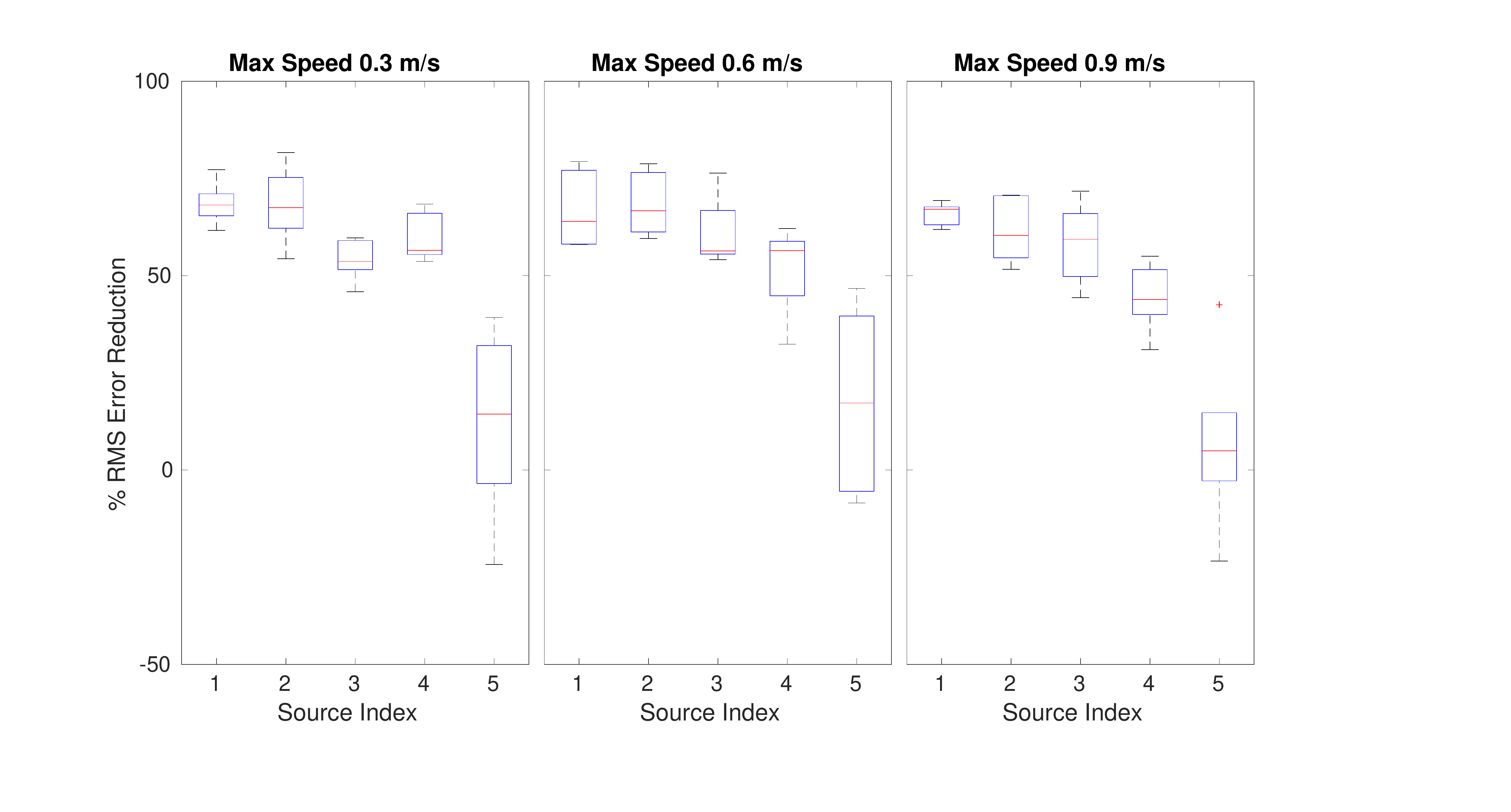}
	\caption{A summary of error reduction as compared to the baseline controller using the inverse dynamics model from each source system at different operating speeds. Using an inverse dynamics model from quadrotors deemed more similar by the proposed metric (Fig.~\ref{fig:gap-plot-experiments}) yields better tracking performance, while using the inverse from a less similar quadrotor (source 5) can even result in \textit{worse} performance as compared to the baseline controller.} 
	\label{fig:error-reduction-experiments}
\end{figure}

\begin{table}[t]
    \centering
    \caption{Summary of Transfer Performance Over 15 Trajectories}
    \begin{tabular}{c|c|c|c|c|c}
    \hline\hline
        & Source 1 & Source 2 & Source 3 & Source 4 & Source 5 \\ 
       ($\hat{\psi}_n^*$) & (0.07) & (0.19) & (0.44) & (0.48) & (0.52) \\
        \hline
        $e_\text{rms}$ [m] & 0.09 & 0.10 & 0.12 & 0.14 & 0.25 \\
        \% Red. & 67.2\% & 66.2\% & 57.9\% & 52.1\% & 12.3\%\\
    \hline\hline
    \end{tabular}
    \raggedright
    \footnotesize{* $e_\text{rms}$ denotes the average RMS tracking error of the target robot with different source experience, and `\% Red.' is the average tracking error reduction as compared to the baseline performance.}
    \label{tab:error-reduction}
\end{table}

We select the input and output of the inverse dynamics module based on~\cite{zhou-cdc17}, and fit the GP with the latest $50$ data points at each timestep. We evaluate tracking performance on 15 hand-drawn trajectories in the $x$-$y$ plane by scaling the drawings in Fig.~\ref{fig:desired_trajectory} based on three maximum speeds $\{0.3, 0.6, 0.9\}$~m/s. We use the root-mean-square (RMS) tracking error as the performance metric.
%

Figure~\ref{fig:error-reduction-experiments} summarizes the performance of the target quadrotor with inverse dynamics modules transferred from the five different source quadrotors (Fig.~\ref{fig:block-diagram}) on the fifteen test trajectories. We note a general correlation between the proposed similarity metric and performance: using the inverse dynamics module from more similar robots results in improved performance. Table~\ref{tab:error-reduction} summarizes the performance of the target robot over the 15 test trajectories when the experience from different source quadrotors is used, again demonstrating that source experience selected by the proposed similarity metric leads to improved transfer performance. Using the experience from the most similar source quadrotor ($R_{S_1}$) yields 62\% better performance than using the experience from an otherwise poorly selected source quadrotor ($R_{S_5}$). These results confirm that the proposed experience selection algorithm based on dynamics similarity is able to predict relative transfer performance with multiple source robots.


\section{Discussion}
\label{sec:discussion}
In this work, we showed that the proposed experience selection algorithm can inform the \textit{relative} transfer performance with different source robots. However, in practice, it would be beneficial to have a threshold on robot similarity that will guarantee positive transfer. One may show that the $\nu$-gap metric is linked to a lower bound on the tracking error for linear systems. A valuable extension is providing theoretical guarantees on the transfer performance of the target robot. Moreover, we note that the $\nu$-gap is developed for linear dynamical systems. While we showed the proposed similarity characterization effectively improves quadrotor tracking performance, we would like to explore experience selection approaches for more challenging nonlinear robot~systems.
\section{Conclusion}
\label{sec:conclusions}
Inspired by the $\nu$-gap metric from robust control theory, we introduced an experience selection algorithm using a characterization of dynamics similarity between robots that can be determined through simple experiments on periodic trajectories. Using this characterization of similarity, we proposed a BO-based algorithm for experience selection when the dynamics of the source and target robots are unknown. We showed in experiments that transferring an inverse dynamics module selected based on the proposed algorithm yields 62\% better performance than an otherwise inappropriately selected experience, illustrating the importance of dynamics similarity in knowledge transfer.

\balance
\bibliographystyle{IEEEtran}
\bibliography{IEEEabrv,reference}


\end{document}